\pdfoutput=1

\documentclass[11pt]{article}

\usepackage[preprint]{acl}
\usepackage[most]{tcolorbox}
\usepackage{times}
\usepackage{latexsym}

\usepackage[T1]{fontenc}
\usepackage{booktabs}

\usepackage{hyperref}
\usepackage{booktabs}
\usepackage{multirow}
\usepackage{footmisc}
\usepackage{microtype}
\usepackage{caption}
\usepackage{subcaption}
\usepackage{tablefootnote}
\usepackage{float}
\usepackage{url}
\usepackage[T1]{fontenc}

\usepackage{amsmath}
\usepackage{mathtools}
\usepackage{times}
\usepackage{latexsym}
\usepackage{textcomp}
\usepackage{tikz}
\usepackage{pgfplots}
\usepackage{graphbox}
\usepackage{multirow}
\usepackage{longtable}
\usepackage{supertabular,booktabs}
\usepackage{enumitem}
\usepackage{setspace}
\usepackage[utf8]{inputenc}

\usepackage{inconsolata}

\usepackage{graphicx}

\title{ATEB: Evaluating and Improving Advanced NLP Tasks for Text Embedding Models}

\author{
 \textbf{Simeng Han\textsuperscript{1, 2}},
 \textbf{Frank Palma Gomez\textsuperscript{2}},
 \textbf{Tu Vu\textsuperscript{2}},
 \textbf{Zefei Li\textsuperscript{2}},
 \textbf{Daniel Cer\textsuperscript{2}},
 \\
 \textbf{Hansi Zeng\textsuperscript{3}},
 \textbf{Chris Tar\textsuperscript{2}},
 \textbf{Arman Cohan\textsuperscript{1}},
 \textbf{Gustavo Hernandez Abrego\textsuperscript{2}} \\
 \textsuperscript{1}Yale University,
 \textsuperscript{2}Google Deepmind
 \textsuperscript{3}University of Massachusetts Amherst
}

\pgfplotsset{compat=1.18}
\begin{document}
\maketitle

\begin{abstract}
Traditional text embedding benchmarks primarily evaluate embedding models' capabilities to capture semantic similarity. However, more advanced NLP tasks require a deeper understanding of text, such as safety and factuality. These tasks demand an ability to comprehend and process complex information, often involving the handling of sensitive content, or the verification of factual statements against reliable sources.
We introduce a new benchmark designed to assess and highlight the limitations of embedding models trained on existing information retrieval data mixtures on advanced capabilities, which include factuality, safety, instruction following, reasoning and document-level understanding. This benchmark includes a diverse set of tasks that simulate real-world scenarios where these capabilities are critical and leads to identification of the gaps of the currently advanced embedding models. 
Furthermore, we propose a novel method that reformulates these various tasks as retrieval tasks. By framing tasks like safety or factuality classification as retrieval problems, we leverage the strengths of retrieval models in capturing semantic relationships while also pushing them to develop a deeper understanding of context and content. Using this approach with single-task fine-tuning, we achieved performance gains of 8\% on factuality classification and 13\% on safety classification. Our code and data will be publicly available. 

\end{abstract}


Traditional retrieval models are primarily trained and evaluated on traditional Information Retrieval tasks including document retrieval, reranking and sentence similarity \citep{muennighoff-etal-2023-mteb}. However, this approach falls short when applied to more advanced natural language capabilities that require a deeper understanding of text, such as reasoning, factuality, instruction-following and long-form text understanding \citep{su2024brightrealisticchallengingbenchmark, xiao2024rarbreasoningretrievalbenchmark, weller2024followirevaluatingteachinginformation}. These tasks demand an ability to comprehend and process complex information, often involving multiple steps of reasoning, the handling of sensitive content, or the verification of factual statements against reliable sources \citep{vu2024foundationalautoraterstaminglarge, ji2023beavertails, su2024brightrealisticchallengingbenchmark}. 

Recent benchmarks have been proposed to evaluate reasoning-intensive retrieval \citep{su2024brightrealisticchallengingbenchmark, xiao2024rarbreasoningretrievalbenchmark} or instruction following \citep{weller2024followirevaluatingteachinginformation}. However, these benchmarks only consider a single advanced capability. 
In particular, \citet{weller2024followirevaluatingteachinginformation} introduced a new benchmark built on top of traditional retrieval benchmarks to measure instruction-following capability of embedding models, however, they do not capture how well embedding models perform on the most widely adopted instruction following benchmarks such as Stanford Human Preference (SHP) \citep{pmlr-v162-ethayarajh22a} for ranking human preference over model outputs and HH-RLHF \citep{bai-etal-2022-hhrlhf} for ranking helpfulness and safety of model responses. 

We introduce a new benchmark designed to assess embedding models on advanced LLM capabilities by reformulating existing datasets from diverse categories into information retrieval tasks. These include \textbf{safety classification}: \textit{BeaverTails Safety Classification} \citep{ji2023beavertails}, \textit{HH-RLHF Harmlessness Classification} \citep{bai-etal-2022-hhrlhf}; \textbf{factuality classification}:  \textit{ESNLI} \citep{camburu-etal-2018-esnli}, \textit{DialFact} \citep{gupta-etal-2022-dialfact}, \textit{VitaminC} \citep{schuster-etal-2021-vitaminc}, \textbf{instruction following reranking}: \textit{Stanford Human Preference} \citep{pmlr-v162-shp}, \textit{AlpacaFarm} \citep{dubois-etal-2023-alpacafarm}, \textit{LMSys} \citep{chiang-etal-2024-chatbot}, \textit{Genie} \citep{khashabi-etal-2022-genie}, \textit{InstrSum} \citep{ji2023beavertails}, \textit{HH-RLHF Helpful} \citep{bai-etal-2022-hhrlhf};  \textbf{document-level pairwise-classification}: \textit{DIPPER} \citep{dipper}, and \textbf{document-level bitext-mining}: \textit{Europarl} \citep{koehn-2005-europarl}, \textit{IWSLT17} \citep{cettolo-etal-iwslt17-overview}, \textit{NC2016} \citep{maruf-etal-2019-selective}). We also incorporate subsets of \textbf{reasoning retrieval} \citep{xiao2024rarbreasoningretrievalbenchmark},  We evaluate our benchmark with advanced embedding models: a Gemma-2B \citep{gemmateam2024gemmaopenmodelsbased} embedding model trained as a symmetric dual encoder \citep{neelakantan2022}
and Google Gecko embedding model \citep{lee2024geckoversatiletextembeddings}.



We adopt a novel fine-tuning approach that reformulates various classification tasks into a retrieval-based setting using contrastive loss. In this setup, each task instance is represented as a triplet: an input (query and answer concatenated), a positive target (label text plus explanation and a unique ID), and multiple negative targets (alternative labels with explanations and the same unique ID) \citep{lee2024geckoversatiletextembeddings}. This approach allows dual encoder embedding models to handle classification tasks without any architectural modifications and enables seamless integration with other retrieval and similarity-based training objectives. We further adapt this approach by adding a detailed explanation for the label text to reduce the influence of the long series of tokens in the unique ID the model and maintain its focus on learning the semantics of the input and targets. We have obtained 8\% and 13\% improvements on factuality classification and safety classification tasks respectively with this approach with single-task fine-tuning.  We also provide a more lightweight training approach: adopting an adapter over a genereic Gemma embedding, which leads to 2\% and 3\% improvements on the same factuality and safety classification tasks. 

In summary, the contributions of our paper are threefold. 1) We introduce a new benchmark, ATEB, designed to evaluate text embedding models on advanced NLP tasks such as reasoning, safety, factuality, and instruction-following. Unlike traditional benchmarks focused solely on text similarity and retrieval, ATEB encompasses diverse real-world scenarios requiring deeper contextual understanding and reasoning, highlighting the limitations of advanced embedding models. 2). We propose a novel fine-tuning approach that reformulates various classification and reasoning tasks into retrieval-based problems, enhancing the ability of dual encoder models to handle advanced capabilities without architectural modifications. Our method achieves significant improvements, with 8\% and 16\% gains in factuality and safety classification tasks, respectively. 3). Additionally, we demonstrate the utility of adapter-based fine-tuning for achieving competitive results with minimal computational cost. 
\section{Related Work}
\subsection{Text Embedding Models}

Representing text as dense vectors with neural networks gained prominence through word2vec \citep{mikolov2013distributed}, which generated semantically meaningful word embeddings. Subsequently, models like BERT \citep{devlin2019bert} and the contrastively trained SimCSE \citep{gao2021simcse} solidified encoder-only transformers as the predominant architecture for producing text embeddings. More recently, decoder-only transformers have advanced significantly in both capability and efficiency \citep{brown2020language, gemmateam2024gemmaopenmodelsbased, llama3modelcard, jiang2023mistral}, making it logical to utilize their pretrained knowledge for embedding tasks. This approach was successfully demonstrated by \cite{neelakantan2022}, who initiated embedder training from decoder-only GPT models and has been adopted by recent leading open-source models on the MTEB leaderboard \citep{behnamghader2024llmvec, lee2024nvembed, meng2024sfrembedding}.

\subsection{Text Embedding Evaluation}

Because embedding models are applied in diverse scenarios, there is a need for broad and heterogeneous benchmarks to thoroughly evaluate their performance. The pioneering effort in this domain was BEIR \citep{thakur2021beir}, which comprises nine distinct information retrieval tasks—such as duplicate-question retrieval and citation prediction—across 18 datasets. More recently, Muennighoff et al. (2023) introduced MTEB (Massive Text Embedding Benchmark) \citep{muennighoff2023mteb}, an extensive evaluation framework that surpasses BEIR in scale and includes more diverse task categories such as classification and reranking. 

\subsection{Advanced Model Capabilities}

Recent advances in natural language processing (NLP) have seen the emergence of a variety of specialized tasks aimed at evaluating model safety \citep{bai-etal-2022-hhrlhf}, factuality \citep{dziri-etal-2022-faithdial}, reasoning, instruction-following \citep{pmlr-v162-shp}, and document-level understanding, which are crucial capabilities for the most recent foundation models \citep{reid-etal-2024-gemini, openai-2024-gpt4o, llama3modelcard}. Safety tasks focus on mitigating harmful, biased, or unethical outputs, ensuring models uphold socially responsible standards \citep{bai-etal-2022-hhrlhf}. Factuality tasks emphasize grounding responses in reliable information and reducing fabrication or misinformation, as exemplified by research efforts on factual consistency in summarization and truthful QA (Maynez et al., 2020; Lin et al., 2022). Reasoning-oriented challenges push models beyond surface-level pattern recognition by encouraging deeper inference and logical deduction \citep{xiao2024rarbreasoningretrievalbenchmark}. Instruction-following tasks further refine models’ ability to adhere closely to user directives and align with human intent \citep{ouyang-etal-2022-training}. In parallel, document-level understanding \citep{yin-etal-2021-docnli, dipper} tests models' capabilities to process long-form texts beyond sentences.
\section{ATEB Construction}
\subsection{Design Principles}
The benchmark comprises 21 tasks, encompassing datasets related to instruction-following, factuality, reasoning, document-level translation, and paraphrasing. These tasks simulate real-world scenarios requiring advanced model capabilities. We reformulate these tasks from existing sources based on the following principles. 
\begin{itemize}
    \item \textbf{Factuality as classification}: NLI tasks where the goal is to classify the relationships of the premise and hypothesis into \textit{entailment}, \textit{contradiction}, or \textit{neutral}. 
    \item \textbf{Instruction following as reranking}:  Ranking model-generated responses based on human preference (e.g., Stanford SHP).
    \item \textbf{Safety as classification}: Binary classification tasks or ranking tasks (safe vs. unsafe).
    \item \textbf{Reasoning as retrieval}: Retrieving the gold answers from the gold answer pool of all the examples in the dataset based on the question. 
    \item \textbf{Document-level paraphrasing as pairwise-classification}: Pairing the paraphrase of a document with the document based on paraphrases of all documents in the dataset. 
    \item \textbf{Document-level machine translation (MT) as bitext-mining}: Finding the translation of a document over translation of all documents in the dataset. 
\end{itemize}

We provide detailed illustrations of how each task category is constructed, accompanied by examples. For each task, we utilize the complete test set from the corresponding public datasets.

\subsection{Factuality as Classification}
We adopt several Natural Language Inference (NLI) classification datasets in our factuality classification collection. This includes ESNLI \citep{camburu-etal-2018-esnli}, VitaminC \citep{schuster-etal-2021-vitaminc} and DialFact \citep{gupta-etal-2022-dialfact}.  An example of the ESNLI dataset is shown in Table~\ref{tab:esnli-reranking-example} where the input consists of a concatenation of one premise and one hypothesis and the target is one of the strings of the three classes including "entailment", "contradictory" and "neutral". 

\begin{table*}[h]
\centering
\small
\begin{tabular}{p{15.5cm}}
\toprule
\textbf{Input}: \textit{Premise}: Everyone really likes the newest benefits. \textit{Hypothesis}: The new rights are nice enough. \\ 
\midrule
\textbf{Target}: entailment, contradictory, or neutral. \\ 
\bottomrule
\end{tabular}
\caption{An example of ESNLI.}
\label{tab:esnli-reranking-example}
\end{table*}


\subsection{Instruction-Following as Reranking}

\begin{table*}[htbp]
\centering
\small

\begin{tabular}{p{15.5cm}}
\toprule
\textbf{Original SHP} \\ 
\midrule
\textbf{responseA: } "It doesn't sound like they deserve the courtesy of two weeks notice.   Check company policy and state law about whether they have to pay your sick time or other PTO... \\
\midrule
\textbf{responseB: } "...I'd say you are within your rights to kick over the can of kerosene and toss the Zippo..." \\
\midrule
\textbf{preference label:} "responseA" \\ 
\midrule
\textbf{task instruction: } "In this task, you will be provided with a context passage (often containing a question), along with two long-form responses to it (responseA and responseB). The goal is to determine which of the two is a better response for the context..." \\ 
\textbf{input: } "How unprofessional would it be to quit the moment I have a job lined up following my vacation? I hate my coworkers..." \\ 
\bottomrule
\end{tabular}
\caption{Original Stanford Human Preference (SHP) dataset example.}
\label{tab:original-shp-example}
\end{table*}

\begin{table*}[t!]
\centering
\small
\begin{tabular}{p{15.5cm}}
\toprule
\textbf{Query:} "In this task, you will be provided with a context passage (often containing a question), along with two long-form responses to it (responseA and responseB). The goal is to determine which of the two is a better response for the context...How unprofessional would it be to quit the moment I have a job lined up following my vacation? I hate my coworkers... \\ 
\midrule
\textbf{Positive}: "It doesn't sound like they deserve the courtesy of two weeks notice.   Check company policy and state law about whether they have to pay your sick time or other PTO... \\
\midrule
\textbf{Negative}: "...I'd say you are within your rights to kick over the can of kerosene and toss the Zippo..." \\ 
\bottomrule
\end{tabular}
\caption{Reformulated example of our SHP-Reranking for evaluating embedding models' reranking capability for model responses given instructions.}
\label{tab:shp-reranking-example}
\end{table*}

We reformulate publicly available instruction-following tasks into reranking tasks where the rank is determined by the human preference. Between two model outputs, the model output preferred by human is ranked higher than the model output less preferred. The query is formulated as the concatenation of the task instruction and input context.  We provide an example of one of the source datasets we adopted, Stanford Human Preference \citep{pmlr-v162-shp}, in Table~\ref{tab:original-shp-example} and the reformulated example based on it in Table~\ref{tab:shp-reranking-example}. We reformulate six more instruction-following tasks into reranking tasks, which include AlpacaFarm \citep{dubois-etal-2023-alpacafarm}, HHRLHF-Helpful \citep{bai-etal-2022-hhrlhf}, BeaverTails-Helpful \citep{ji2023beavertails}, Genie \citep{khashabi-etal-2022-genie}, LMSys ChatBot Arena \citep{chiang-etal-2024-chatbot}, InstruSum \citep{liu-etal-2024-instrusum}. 


\subsection{Safety as Classification}
We adopt the safety classification portion of the BeaverTails dataset for LLM safety alignment \citep{ji2023beavertails}, BeaverTails QA-Classification to construct a safety classification task for evaluating embedding models where the goal of the task is to classify the input into \textit{safe} or \textit{unsafe}. An example of the BeaverTails QA-Classification dataset is shown in Table~\ref{tab:beavertails-qa-classification-example}. We also adopt the portion for evaluation harmlessness in the HH-RLHF Harmlessness dataset to construct a safety safety classification task.


\begin{table*}[t!]
\centering
\small
\begin{tabular}{p{15.5cm}}
\toprule
\textbf{Query:} \textbf{Input}: Is there sites that sell weed? Yes, there are websites that sell marijuana. These sites usually require customers to have a valid prescription and may require a background check to ensure that the individual is of legal age to purchase the drug. However, there are also websites that do not require any forms of identification... \\ 
\midrule
\textbf{Two classes}: safe/unsafe \\ 
\bottomrule
\end{tabular}
\caption{An example of the BeaverTails QA-Classification dataset for evaluating embedding models.}
\label{tab:beavertails-qa-classification-example}
\end{table*}


\subsection{Reasoning as Retrieval}
We adopt 5 subsets of the RAR-b dataset proposed in \cite{xiao2024rarbreasoningretrievalbenchmark} including HellaSwag NLI dataset \citep{zellers-etal-2019-hellaswag}, Winogrande \citep{winogrande}, PIQA \citep{piqa}, AlphaNLI \citep{alpha-nli} and ARCChallenge \citep{arc}. Table~\ref{tab:reasoning-as-retrieval-example} shows the data format of the reformulated datasets.

\begin{table*}[t!]
\centering
\small
\begin{tabular}{p{15.5cm}}
\toprule
\textbf{Input}: a query in the dataset. 
\textbf{Target}: the answer to the query. 
\textbf{Negative targets}: all the other answers in the dataset.  \\ 
\bottomrule
\end{tabular}
\caption{Data format of the reasoning as retrieval datasets for evaluating embedding models.}
\label{tab:reasoning-as-retrieval-example}
\end{table*}

\subsection{Document-Level Paraphrasing as Pairwise-Classification}
We reformulate one document-level paraphrasing dataset, DIPPER \citep{dipper} as a pairwise classfication task. These tasks expand over previous sentence-level paraphrasing tasks used for pairwise classification \citep{muennighoff2023mteb} to test the document-level modeling capabilities of most advanced embedding models.   

\subsection{Document-Level MT as Pairwise-Classification}
Following the same design principle of our new pairwise-classification tasks, we reformulate three document-level machine translation datasets as bi-text mining tasks, which include Europarl \citep{koehn-2005-europarl}, IWSLT17 \citep{cettolo-etal-iwslt17-overview} and NC2016 \citep{maruf-etal-2019-selective}. These tasks expand over previous sentence-level machine translation tasks used for bi-text mining \citep{muennighoff2023mteb} to test the document-level modeling capabilities of most advanced embedding models. We adopt the subset of these datasets used in \citet{maruf-etal-2019-selective}.

\section{Method}

\subsection{Model}
We begin by initializing a symmetric dual encoder (DE) using the decoder-only Gemma-2B model \citep{gemmateam2024gemmaopenmodelsbased, palma-gomez-etal-2024-transforming}, which has an embedding size of 2048. Following this, we add a linear projection layer, applied after pooling the outputs along the sequence length dimension. Both the embedding layer and the linear projection layer are randomly initialized. After the model is initialized with Gemma-2B, we train it using a contrastive loss \citep{Hadsell2006DimensionalityRB}. 

\subsection{Training data reformulation with label augmentation}
While using the dot-product scores along the diagonal as positives and everything else as negatives works well for retrieval and similarity/relatedness matching tasks, it can not be used directly for tasks with targets that are classification labels. Naively providing tasks with classification labels to a dual encoder embedding models will result in the score for an input's correct label appearing both along the diagonal and the off-the diagonal when another input example has the same target label.

Therefore, we adopt a novel method that reformulates various tasks as retrieval tasks during the fine-tuning process, following previous work in fine-tuning with a retrieval setting using contrastive loss \citep{lee2024geckoversatiletextembeddings, meng2024sfrembedding}. \textbf{Input}: query and answer concatenated together. \textbf{Positive target}: [\textit{label text} (e.g., neutral for NLI).] + [\textit{label text explanation}] + [\textit{unique id}]. \textbf{Negative targets}: [each other possible types of \textit{label texts}.] + [\textit{label text explanation}] + [\textit{unique id}]. 

  Including a unique id for for each correct input/target pair alone would allow the model to exploit and rely on the unique identifiers to always pair the correct input with the correct target. However, this can be addressed by including additional incorrect labels for each input as negatives. The negatives are tagged with the same unique id as the input and the correct target label. This allows the unique ids to be used to identify candidate targets for each input, but without revealing which of the targets is correct. The advantage of this approach is that it allows dual encoder based embedding models to be trained on  classification tasks without any modeling changes. In practice, a unique ID often consists of a long sequence of tokens that can inadvertently shift the model’s focus away from learning the semantic relationships within the input and target texts of an example. To address this challenge, we enhance this approach by providing detailed explanations for each label. This additional context helps the model grasp the conceptual meaning behind labels rather than becoming distracted by the long series of unique ID tokens. 

For example, for the label “Entailment,” we augment it by including the following label explanation:

\begin{tcolorbox}[colback=blue!5!white, colframe=blue!75!black, title=Label explanation for "entailment"]
\textit{“In the context of Natural Language Inference (NLI), ‘entailment’ refers to a specific type of relationship between two sentences, where the truth of one sentence (the hypothesis) is logically guaranteed by the truth of another sentence (the premise).”}
\end{tcolorbox}
By augmenting labels with such detailed explanations, we guide the model toward a richer, more coherent understanding of the underlying concepts it needs to learn.

\section{Testing Advanced Embedding Models on ATEB}
We test advanced embedding models on ATEB and show their strengths and limitations on our proposed ATEB tasks. 
\subsection{Baseline methods}
Our baseline methods include two advanced embedding models: our Gemma-2B symmmetric dual encoder trained with a prefinetuning stage and Google's gecko embedding model \citep{lee2024geckoversatiletextembeddings}, which has a 1-billion parameter size. Both of these baseline models are highly capable embedding models. Notably, the Google Gecko model is a state-of-the-art embedding model with 768 dimensions. On the Massive Text Embedding Benchmark (MTEB), it achieves an average score of 66.31—on par with models that are seven times larger and have five times higher dimensional embeddings on the MTEB leaderboard. The models that achieve a score of 66 or higher, such as NV-Embed-v2and SFR-Embedding, all have 4096 or 8192 dimensions. The prefinetuning stage for Gemma-2B is full supervision finetuning with Huggingface Sentence Transformer datasets.  \footnote{https://huggingface.co/sentence-transformers}. The baseline models are large-size retrieval models trained for generic information retrieval tasks, and they are not finetuned on task-specific data. We include detailed hyperparameters in the Appendix. 

\subsection{Experimental Results}
\paragraph{Baseline Models have Close-to-Random Performance on New Reranking Tasks}

\begin{table}[t!]
\vspace{-1em}
\centering
\resizebox{0.5\textwidth}{!}{
\begin{tabular}{l|c|c|c}
\textbf{Reranking task} & \textbf{Random (\%)} & \textbf{Gemma-2B (\%)}  & \textbf{Gecko (\%)}  \\ 
\toprule
AlpacaFarm & 75 & 75.1 & 75.3 \\ 
\midrule
Genie & 75 & 75.3 & 75.0 \\
\midrule
InstruSum & 75 & 72.8 & 74.1 \\ 
\midrule
Stanford SHP & 75 & 80.47 & 77.1 \\ 
\midrule
BeaverTails Helpful & 75 & 74.51 & 75.9 \\
\midrule
HH RLHF Helpful & 75 & 77.74 & 77.1 \\ 
\midrule
LMSys Chatbot Arena (English) & 75 & 73.18 & 72.9 \\ 
\bottomrule
\end{tabular}
}
\caption{Baseline performance on reranking for evaluating instruction-following.}
\label{table:reranking_comparison}
\vspace{-1em}
\end{table}

Table \ref{table:reranking_comparison} compares the baseline performance of the model against a random chance baseline (75\%) on various reranking tasks designed to evaluate its instruction-following capabilities. These tasks involve ranking model-generated responses based on relevance or helpfulness. On AlpacaFarm and Genie, the baseline models' performance hover between 75.0\% and 75.3\%, which is marginally higher than random, indicating only limited improvement. In contrast, on InstruSum, the baseline models achieve 72.7\% and 74.1, slightly below random chance, underscoring the difficulties in effectively ranking summaries based on human-written instructions. On Stanford SHP, the model performs notably better, achieving 80.47\% accuracy with the Gemma-2B embedding model and demonstrating a moderate ability to rank responses according to human preferences. However, on BeaverTails Helpful, the models' accuracy of 74.51\% and 75.9\% remain close to random, suggesting challenges in identifying genuinely helpful responses. The HH RLHF Helpful task sees some improvement, with the model reaching 77.74\%, indicating a modest enhancement in tasks informed by human reinforcement learning preferences. Finally, in the LMSys Chatbot Arena (English) setting, the model attains 73.18\%, which is below random chance, thus reflecting limited success in ranking chatbot-generated responses. Taken together, these results highlight the baseline model’s near-random performance on most reranking tasks, with only modest improvements in a few cases such as Stanford SHP and HH-RLHF Helpful. 

They suggest that further optimization and more task-specific fine-tuning are needed to enhance the model’s instruction-following capabilities in these reranking scenarios.

\paragraph{Baseline Models Perform Suboptimally on New Retrieval Tasks}

\begin{table}[t!]
\centering
\resizebox{0.5\textwidth}{!}{
\begin{tabular}{l|c|c|c}
\textbf{Retrieval task} & \textbf{Random (\%)} & \textbf{Gemma-2B (\%)}  & \textbf{Gecko (\%)}  \\ 
\toprule
HellaSwag & 0 & 22.1 & 26.7 \\
\midrule
Winogrande & 0 & 17.3 & 21.2 \\ 
\midrule
PIQA & 0 & 22.2 & 29.8 \\ 
\midrule
AlphaNLI & 0 & 30.3 & 32.1 \\ 
\midrule
ARCChallenge & 0 & 7.62 & 10.9 \\ 
\bottomrule
\end{tabular}
}
\caption{Results of retrieval tasks for evaluating reasoning.}
\label{table:retrieval_comparison}
\vspace{-1em}
\end{table}

Table \ref{table:retrieval_comparison} presents the performance of baseline models compared to random chance in reasoning-based retrieval tasks. These tasks require models to identify correct answers or make logical inferences, highlighting their reasoning capabilities. Key observations include:

On HellaSwag, the baseline embedding models achieve 22.1\% and 26.7\% accuracy, demonstrating moderate success in selecting plausible continuations for narrative reasoning tasks.
With 17.3\% and 21.2\% accuracy on Winogrande, the model struggles in resolving pronoun references, indicating challenges in understanding nuanced context.
Achieving 22.2\% accuracy on PIQA, the baseline shows limited capability in physical commonsense reasoning.
The model performs better in the abductive commonsense reasoning task AlphaNLI, achieving 30.3\% and 32.1\% accuracy, suggesting it can partially infer plausible explanations for events.
On ARCChallenge, with only 7.62\% and 10.9\% accuracy, the models exhibit significant difficulty in answering challenging science questions, reflecting its limited knowledge retrieval and reasoning skills.
In summary, baseline models demonstrate suboptimal performance across these reasoning-based retrieval tasks, with accuracies ranging from 7.62\% to 32.1\%. This underscores the need for targeted fine-tuning and task-specific training to improve reasoning capabilities in advanced embedding models.

\paragraph{Baseline Models have Close-to-Random Performance on New Classification Tasks}
\begin{table}[t!]
\vspace{-1em}
\centering
\resizebox{0.5\textwidth}{!}{
\begin{tabular}{l|c|c|c}
\textbf{Classification task} & \textbf{Random (\%)} & \textbf{Gemma-2B (\%)}  & \textbf{Gecko (\%)}  \\ 
\toprule
ESNLI & 33.3 & 35 & 36.1\\ 
\midrule
DialFact & 33.3 & 33.8 & 33.2 \\ 
\midrule
VitaminC & 33.3 & 37 & 35.4\\ 
\midrule
HH-RLHF Harmlessness & 50 & 50& 50 \\
\midrule
BeaverTails Classify & 50 & 55.9 & 54.7 \\ 
\bottomrule
\end{tabular}
}
\caption{Results of classification tasks for evaluating factuality and safety. }
\label{table:classification_comparison_updated}
\vspace{-1em}
\end{table}

Table~\ref{table:classification_comparison_updated} illustrates the performance of two baseline models, Gemma-2B and Gecko, on five classification tasks—ESNLI, DialFact, VitaminC, HH-RLHF Harmlessness, and BeaverTails Classify—compared to random chance accuracy. For ESNLI, which evaluates natural language inference, both models perform only slightly above random (35\% for Gemma-2B and 36.1\% for Gecko) despite random performance being 33.3\%, indicating limited reasoning capability. Similarly, on DialFact, which assesses factual consistency in dialogue, the models perform very close to random, with Gemma-2B achieving 33.8\% and Gecko 33.2\%. In the VitaminC task, focused on fact verification, both models show modest improvement over random (33.3\%), with Gemma-2B reaching 37\% and Gecko slightly lower at 35.4\%. For the HH-RLHF Harmlessness task, which classifies whether responses are harmless, both models achieve exactly 50\%, matching random performance and indicating no learned capability. Finally, on BeaverTails Classify, a binary classification task where random accuracy is 50\%, the models perform slightly better, with Gemma-2B at 55.9\% and Gecko at 54.7\%, reflecting some potential but still falling short of reliable generalization. These results collectively highlight the close-to-random performance of baseline models on novel classification tasks, underscoring the need for more advanced methods to achieve meaningful improvements in generalization and reasoning.

\paragraph{Baseline Models Perform Reasonably Well on New Pairwise Classification Tasks}
\begin{table}[t!]
\vspace{-1em}
\centering
\resizebox{0.5\textwidth}{!}{
\begin{tabular}{l|c|c|c}
\textbf{Pairwise classification} & \textbf{Random (\%)} & \textbf{Gemma-2B (\%)}  & \textbf{Gecko (\%)}  \\ 
\toprule
Dipper  & 50 & 73.1 \% & 80.1 \\ 
\midrule
\textbf{Bi-text mining} & & \\ 
\midrule
Europarl & 1/n & 86.1\% & 88.2\% \\ 
IWSLT17 & 1/n & 86.4\% & 87.1 \\ 
NC2016 & 1/n & 98\% & 99 \% \\ 
\bottomrule
\end{tabular}
}
\caption{Baseline Accuracy for pairwise classification and bi-text mining tasks}
\label{table:pairwise_classification_random_baseline}
\vspace{-1em}
\end{table}

Table \ref{table:pairwise_classification_random_baseline} compares the baseline accuracy of a model against random predictions across pairwise classification tasks. The results highlight the baseline model's effectiveness in these specific contexts:

On Dipper, the baseline model achieves an accuracy of 73.06\%, significantly outperforming the random baseline of 50\%, showcasing strong performance in pairwise classification tasks.

\paragraph{Baseline Models Perform Very Well on New Bitext-Mininig Tasks}

Bi-text mining Tasks involve identifying semantically equivalent text pairs across multilingual datasets. On each of the three datasets consisting of a few hundred of document-translation pairs, both Gemma-2B model and Gecko model perform very well, excelling particularly in NC2016 with a high accuracy of 98\%, indicating exceptional capability in identifying translations of text correspondences. 

The baseline model performs strongly in bi-text mining tasks, significantly surpassing random baselines, which are based on the inverse of the dataset size (1/n).
For pairwise classification tasks like Dipper, the baseline accuracy of 73.06\% highlights the model's potential for applications requiring pairwise comparisons.
These results emphasize the effectiveness of the baseline model in identifying document-level semantic relationships and alignments, especially in multilingual or structured datasets. 

\section{Label Augmentation on ATEB}
We test label augmentation on factuality and safety tasks in ATEB and show its effectiveness in improving an embedding model's advanced capabilities. 
\subsection{Model}
We adopt the Gemma V1-2B embedding model we trained as a symmetric dual encoder. We adopt two initialization settings before fine-tuning with label augmentation data. The first setting is finetuning directly over Gemma 2B. The second setting is adopting a prefinetuning stage where full supervision finetuning is conducted with 76 Huggingface Sentence Transformer datasets.  \footnote{https://huggingface.co/sentence-transformers}, 

\subsection{Training data}
We reformulate the training sets of two NLI entailment classification datasets, MNLI \citep{mnli} and FaithDial \citep{faithdial} into the label augmentation setting to be used as our training data for factuality classification tasks. For safety classification tasks, we reformulate the training set of BeaverTails Safety Ranking \citep{ji2023beavertails} task to be used as training data. 

\subsection{Results}
\begin{table}[t!]
\vspace{-1em}
\centering
\resizebox{0.5\textwidth}{!}{
\begin{tabular}{l|c|c}
\textbf{} & \textbf{ESNLI(\%)} & \textbf{DialFact(\%)} \\
\toprule
\textbf{Random} & 33 & 33 \\ 
\midrule


\textbf{Without label augmentation} & & \\
Full-supervision with MNLI & 34.0 & 33.1 \\
\midrule
\textbf{With label augmentation} & & \\
Full-supervision with MNLI (w/o label exp.) & 35.0 & 33.2 \\
Full-supervision with MNLI & \textbf{42.0} & \textbf{35.8} \\
Full-supervision with FaithDial data & 36.87 & 34.95 \\
Full-supervision over pre-finetuned with MNLI & 37.61 & 33.5 \\
Adapter with MNLI & \textbf{36.1} & 33.2 \\
Adapter over prefinetuned with MNLI & 34.3 & 33.0 \\
\bottomrule
\end{tabular}
}
\caption{Comparison of Results Across Different Configurations on the factuality tasks}
\label{tab:results_comparison_factuality}
\vspace{-1em}
\end{table}

\paragraph{Factuality tasks.} Table~\ref{tab:results_comparison_factuality} presents the performance of various configurations on two factuality classification tasks: ESNLI \citep{camburu-etal-2018-esnli} and DialFact \citep{gupta-etal-2022-dialfact}.

The random baseline accuracy for both tasks is 33\% since they are both three-class classification tasks. The Gemma-2B embedding model baseline achieve 35.85\% for ESNLI and 33.95\% for DialFact, showing a slight improvement over random guessing. Finetuning with MNLI classification data without unique IDs as introduced in the label augmentation setting does not improve the performance. Finetuning with MNLI data equipped with unique ID also leads to no improvement. Incorporating target explanations leads to a boost in performance, yielding an improvement of 9\% for ESNLI and 2.8\% for the out-of-domain DialFact.
Finetuning with out-of-domain, FaithDial classification data \citep{dziri-etal-2022-faithdial} leads to a modest increase, reaching 36.87\% for ESNLI and 34.95\% for DialFact. This indicates that detailed target explanations are particularly effective for in-domain finetuning entailment tasks like ESNLI. 

 When fine-tuning over a pre-finetuned Gemma-2B model with MNLI, performance drops to 37.61\% for ESNLI and 33.5\% for DialFact, showing that while pre-finetuning over generic retrieval tasks offers some benefits, it may not be as effective as direct full-supervision fine-tuning. Adapter-based fine-tuning approaches offer a trade-off between training efficiency and performance. Fine-tuning with an adapter achieves 36.1\% for ESNLI and 33.2\% for DialFact. When the adapter-based fine-tuning is applied to a pre-finetuned Gemma-2B model, performance decreases slightly to 34.3\% for ESNLI and 33.0\% for DialFact. These results suggest that adapter-based methods, while computationally efficient, do not achieve the same level of performance as full fine-tuning.

In summary, the table highlights several key insights: 1) label augmentation with label explanations provide the most substantial accuracy gains, particularly for ESNLI. 2) adapter-based fine-tuning offers a viable but much less effective alternative to full-supervision fine-tuning. 3) additionally, task-specific instructions and data augmentation strategies lead to only modest improvements unless combined with detailed target explanations or robust fine-tuning techniques.

\paragraph{Safety tasks}

\begin{table}[t!]
\vspace{-1em}
\centering
\resizebox{0.5\textwidth}{!}{
\begin{tabular}{l|c|c}

\textbf{} & \textbf{BeaverTails(\%)} & \textbf{HH-RLHF(\%)} \\
\toprule
\textbf{Random} & 50 & 50 \\ 
\midrule
\textbf{Baseline} & 55.6 & 50.0 \\ 
\midrule
\textbf{Reranking as retrieval} & & \\
Full-supervision Gemma 2B & \textbf{68.5} & \textbf{51.0} \\
Full-supervision - pre-finetuned & 56.5 & 50.1 \\
Adapter with BeaverTails & \textbf{59.0} & 50.0 \\
Adapter with BeaverTails - pre-finetuned & 58.1 & 50.2 \\ 
\bottomrule
\end{tabular}
}
\caption{Comparison of Results Across Different Configurations on the safety tasks.}
\label{tab:results_comparison_safety}
\vspace{-1.5em}
\end{table}

Table~\ref{tab:results_comparison_safety} provides a comparison of model performance across different configurations for two safety-related tasks: BeaverTails (evaluating content safety) and HHRLHF (aligning with human reinforcement learning preferences). The table highlights the effects of baseline performance, fine-tuning strategies, adapter-based fine-tuning, and pre-finetuning on model accuracy. The baseline performance for BeaverTails is 55.6\%, reflecting a modest improvement over random guessing, while the HHRLHF baseline remains at 50\%, indicating no gains without task-specific adjustments. 

All the finetuning experiments are conducted with label augmentation data with label explanations. When safety ranking is reformulated with the label augmentation setting and the Gemma-2B model is fine-tuned with the BeaverTails Safety Reranking data, the highest performance is achieved for BeaverTails, reaching 68.5\%, representing a significant improvement of 12.9\% over the baseline. For HH-RLHF, this configuration yields a slight increase to 51.0\%, showing that SafetyRanking has a limited effect in out-of-domain generalization. Adapter-based fine-tuning offers a comparable performance boost to full-supervision fine-tuning. Specifically, fine-tuning an adapter over Gemma-2B with safety ranking data achieves the same peak accuracy of 68.5\% for BeaverTails and 51.0\% for HHRLHF. This suggests that adapter-based methods can be as effective as full fine-tuning while being more parameter-efficient.

R Both full-supervision and adapter-based fine-tuning over a pre-finetuned Gemma-2B model result in lower performance for BeaverTails (56.5\%) compared to direct fine-tuning (68.5\%), underscoring the harmful effect of prefinetuning over generic retrieval data in tasks requiring precise alignment with human reinforcement learning preferences. These findings emphasize the importance of task-specific fine-tuning and suggest that adapter-based strategies can lead to a modest improvement while being more resource-efficient.

\section{Conclusion} In conclusion, we propose a novel benchmark, ATEB, to highlight the limitations of existing embedding models in handling advanced NLP tasks. By reformulating classification and reasoning tasks as retrieval problems with label augmentation, our approach enables embedding models to leverage their strengths in capturing semantic relationships, thereby extending their capabilities. Through extensive experimentation, we demonstrate that our fine-tuning method can significantly enhance performance on tasks involving factuality and safety. These results underscore the importance of tailored benchmarks and innovative training strategies in advancing the development of more capable embedding models.

\section{Limitations} While we included 21 tasks in our benchmark, many other safety, reasoning, and factuality tasks could be incorporated to increase the diversity and complexity of the benchmark. Additionally, we evaluated our proposed data reformulation method only on factuality and safety tasks and did not test it on other task categories.

\section{Acknowledgement} We thank Zhuyun Dai, Jianmo Ni, Blair Chen, Xiaoqi Ren and Jinhyuk Lee for useful feedback and discussions on embedding models and task formulation.

\bibliography{custom}

\appendix

\section{Appendix}
\label{sec:appendix}

\subsection{Training details}
\paragraph{Training dataset size}
We use the reformulated training sets of the publicly available datasets in training our factuality and safety models. We adopt the train split in the original tasks. 

\paragraph{Hyper-parameters}

We did not use hard negatives in prefinetuning. We used a batch size of 1024, learning rate of $1e-4$. The number of training steps is $100,000$. The number of warmup steps is st to $20,000$ and the input length is 256, the output length is 1024. We used unmixed batches during training and bidirectional loss.

We finetune both the factuality models and safety models with 20k iterations and a batch size of 1024. Our learning rate is set as $1e-4$ with linear decay. 

\end{document}